\documentclass{article}

\usepackage[nonatbib,final]{nips_2018}
\usepackage{tikz}
\usetikzlibrary{bayesnet}
\usetikzlibrary{positioning}
\usepackage{amsmath}
\usepackage{floatrow}

\usepackage{subfig}
\usepackage[utf8]{inputenc} % allow utf-8 input
\usepackage[T1]{fontenc}    % use 8-bit T1 fonts
\usepackage{hyperref}       % hyperlinks
\usepackage{url}            % simple URL typesetting
\usepackage{booktabs}       % professional-quality tables
\usepackage{amsfonts}       % blackboard math symbols
\usepackage{nicefrac}       % compact symbols for 1/2, etc.
\usepackage{microtype}      % microtypography

\title{A Bayesian Solution to the M-Bias Problem}

\author{
  David Rohde\\
  Criteo AI Lab\\
  Paris\\
  \texttt{d.rohde@criteo.com} \\
}

\begin{document}

\maketitle

\begin{abstract}
It is common practice in using regression type models for inferring
causal effects, that inferring the correct causal relationship requires
extra covariates are included or ``adjusted for''.  Without performing this adjustment
erroneous causal effects can be inferred.
Given this phenomenon it is common practice
to include as many covariates as possible, however such advice comes
unstuck in the presence of M-bias. 
M-Bias is a problem in causal inference where the correct estimation
of treatment effects \emph{requires} that certain variables are not
adjusted for i.e. are simply neglected from inclusion in the model.
This issue caused a storm of controversy in 2009 when Rubin, Pearl
and others disagreed about if it could be problematic to include
additional variables in models when inferring causal effects.
This paper makes two contributions to this issue.  Firstly we provide
a Bayesian solution to the M-Bias problem.  The solution replicates
Pearl's solution, but consistent with Rubin's advice we condition on
all variables.  Secondly the fact that we are
able to offer a solution to this problem in Bayesian terms shows that
it is indeed possible to represent causal relationships within the
Bayesian paradigm, albeit in an extended space.  We make several
remarks on the similarities and differences between causal graphical
models which 
implement the do-calculus and probabilistic graphical models which
enable Bayesian statistics.
We hope this work will stimulate more research on
unifying Pearl's causal calculus using causal graphical models with
traditional Bayesian statistics and probabilistic graphical models.
\end{abstract}

\section{Introduction}

In a causal problem we are interested in understanding the outcome $Y$ of
applying treatment $T$ to a user with measured attributes $X$.  It is
well known that if there exist variables that effect
both the treatment assignment $T$ and the outcome $Y$ then these
unobserved effects can confound estimates of the treatment effect, a
phenomenon known as Simpson's Paradox \cite{pearl_simps}.

The usual way to minimize confounding is to attempt to do ``back door
adjustment'' (which in practice usually means including the covariates in a
regression model \cite{pearl1995causal})
for as many observed covariates as are available, this is
despite the risk of M-bias which actually increases if back door
adjustment is applied unthinkingly to all available variables.  An
alternative method for achieving back door adjustment is to use
propensity score methods \cite{rosenblum1983central}.

A storm of controversy started in 2009, when Rubin was asked if there
were ever cases where covariates should not be included in a model 
in the journal \emph{Statistics and
  Medicine} \cite{shrier2009propensity}.  Several discussants responded and a number highlighted that in 
the presence of a specific structure known as the M-structure
adjusting for some variables could  
increase rather than decrease confounding \cite{pearl2009myth}
\cite{sjolander2009propensity} \cite{shrier2009propensity}.

Rubin ultimately stated the standard Bayesian position that all
variables should be conditioned upon, and
removing a variable is an ad hockery \cite{rubin2009should}.

The contribution of this paper is to present a Bayesian solution that
follows Rubin's advice of conditioning on all variables and yet
obtains the solution advocated by Pearl and others, where average
treatment effects unconditional on the covariate can be identified,
but personalized treatment effects conditional on the covariates
cannot be.  In order to allow Bayesian statistics to be applied to
causal problems we introduce a two plates framework for probabilistic
graphical models, where there is a mapping between the pre and post
intervention graphs used by causal graphical models introduced by
Pearl and the two plates in the probabilistic graphical model.  The
two plates framework differentiates itself by having explicit
representation of parameters. The parameters can then in some cases carry information
from the observation plate to the intervention plate allowing the
identification of causal effects.

It is conjectured, but not proven, that the two plates
framework allows probabilistic graphical models to be
identifiable under the same conditions as causal graphical models
\cite{pearl1995causal}, but to have benefits in finite sample problems
or in non-identifiable cases.

In Section \ref{pearlcgm} the Pearl solution is presented,
in Section \ref{twoplatepgm} we provide the Bayesian solution, in
Section \ref{ssmcmc} we use Markov chain Monte Carlo (MCMC) to demonstrate
the method on a case 
study.  We note the posterior reflects that non-identifiability, but
also has some surprising structure.  Concluding remarks are made in
Section \ref{conc}.

\section{The Pearl Causal Graphical Model Solution for M-Bias}
\label{pearlcgm}
\begin{figure}
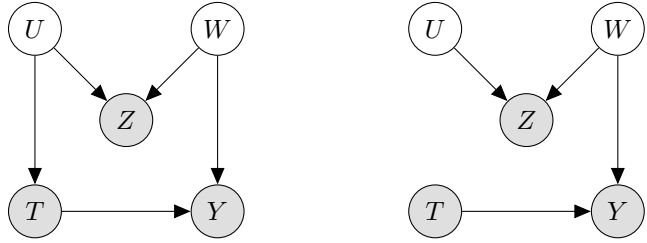

\begin{floatrow}
  \centering
  \tikz{ %
    \node[obs] (Zm) {$Z$} ; %
    \node[obs, below left=of Zm] (Tm) {$T$} ; %
    \node[obs, below right=of Zm] (Ym) {$Y$} ; %
    \node[latent, above right=of Zm] (Wm) {$W$} ; %
    \node[latent, above left=of Zm] (Um) {$U$} ; %
    \edge {Um} {Tm} ; %
    \edge {Um} {Zm} ; %
    \edge {Wm} {Zm} ; %
    \edge {Wm} {Ym} ; %
    \edge {Tm} {Ym} ; %
  }
  \hspace{2cm}
  \tikz{ %
    \node[obs] (Zm) {$Z$} ; %
    \node[obs, below left=of Zm] (Tm) {$T$} ; %
    \node[obs, below right=of Zm] (Ym) {$Y$} ; %
    \node[latent, above right=of Zm] (Wm) {$W$} ; %
    \node[latent, above left=of Zm] (Um) {$U$} ; %
    \edge {Tm} {Ym} ; %
    \edge {Um} {Zm} ; %
    \edge {Wm} {Zm} ; %
    \edge {Wm} {Ym} ; %
  }
  \caption{The M-bias Mutilated Causal Graphical Model: left original
    graph from which the data was collected, right the mutilated graph
    in which we intervene.  Shaded variables are observed,
    clear variables are latent.}
  \label{mbias_cgm}  
\end{floatrow}
\end{figure}

A Causal Graphical Model (CGM) for the M-bias problem is shown in Figure
\ref{mbias_cgm}.  Two
hidden variables must be drawn independently $U$ and $W$,
conditionally on these variables $Z$ is drawn, conditionally on $U$,
$T$ is drawn and conditionally on $W$, finally $Y$ is drawn
conditionally on $W$ and $T$. 

A researcher is observing $T,Z,Y$ from the system given in Figure
\ref{mbias_cgm}. and regresses $Y$ on $T$ will be able to determine
the (average) treatment effect.  However if the researcher follows the 
advice given by some researchers to include all possible covariates
and the variable $Z$ is adjusted for, perhaps surprisingly, erroneous
treatment effects will be found, see \cite{ding2015adjust} for more
discussion. 

The do-calculus can be used to transform a probabilistic specification
as given in Figure \ref{mbias_cgm} (left) to (if it is indeed possible) the mutilated
specification  Figure \ref{mbias_cgm} (right) using three substitution
rules given in \cite{pearl1995causal}. 

In this case we only need the first rule that states that $P(Y|{\rm
  do}(T),Z)$, can be replaced with: $P(y|{\rm do}(T))$, if $Y$ is independent of $Z$
conditional on $T$ in the mutilated graph.  The practical meaning of
this result is that a researcher can ignore the observations of $Z$
and build a model that infers an average treatment effect of $T$ on
$Y$. 

A further curiosity about this problem is that if we observed many
realizations of the mutilated graph Figure \ref{mbias_cgm} (right) it
would be indeed
possible to use $Z$ to produce personalized treatment effects (as $Z$
gives information about $W$ which affects $Y$).  However this
relationship involving $Z$ cannot be identified from the un-mutilated model
Figure \ref{mbias_cgm} (left).

\begin{table}
  \centering
\caption{Example data set.}
\label{data}
\begin{tabular}{|r|r|r|r|}
\hline
T&Z&Y&N\\
\hline
0 & 0 & 0 &33\\
0 & 0 & 1 & 2\\
0 & 1 & 0 & 95\\
0 & 1 & 1 & 50\\
1 & 0 & 0 & 100\\
1 & 0 & 1 & 47\\
1 & 1 & 0 & 60\\
1 & 1 & 1 & 240\\
\hline
\end{tabular}
\end{table}

For purposes of illustration, we assume that $Y, Z, W, T \in \{0, 1\}$ and the data is given in
Table \ref{data}.  According to the do calculus we ignore $Z$ and should compute
estimates of the average treatment effect which is given by:

\[
P(Y=1|T=1) - P(Y=1|T=0)  = \frac{ 47 + 240 } {47 + 240 + 100 + 60 } -
\frac{50 + 2} {50 + 2 + 33 + 95} \approx 0.35
\]

The source of controversy in this example is that it is not
appropriate to apply back door adjustment here.  If we were to apply
back door adjustment we obtain:

\[
P(Y=1|T=1,Z=0) - P(Y=1|T=0,Z=0) = \frac{47} { 47 + 100 }  - \frac{2}
{2+33} \approx 0.26,
\]

\[
P(Y=1|T=1,Z=1) - P(Y=1|T=0,Z=1) = \frac{240} {240 + 60 } - \frac{50}
{50 + 95} \approx 0.46.
\]

This calculation while not causal is valid if you would like to update
your belief about records in the observational data where the $Y$
label is missing.  Crucially it does not apply to the mutilated graph.

\section{The Bayesian Probabilistic Graphical Model Solution for M-Bias}
\label{twoplatepgm}

\begin{figure}
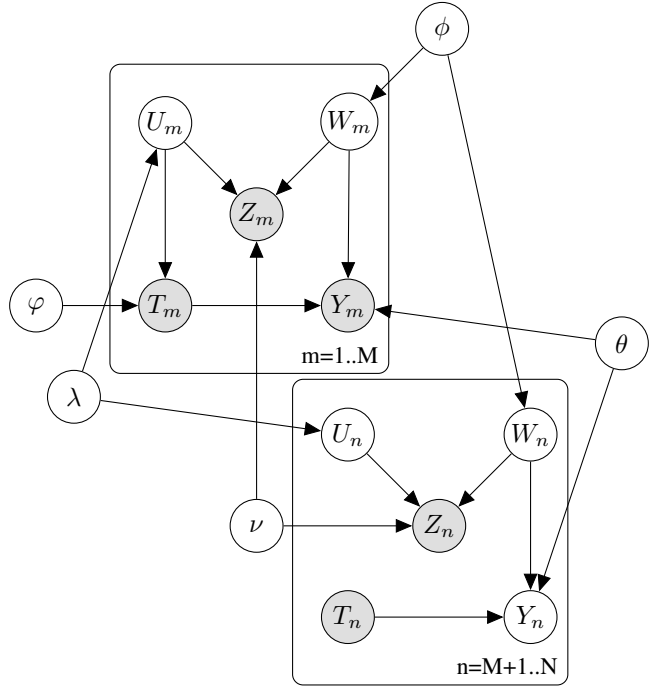

  \centering
  \tikz{ %
    \node[obs] (Zm) {$Z_m$} ; %
    \node[obs, below left=of Zm] (Tm) {$T_m$} ; %
    \node[obs, below right=of Zm] (Ym) {$Y_m$} ; %
    \node[latent, above right=of Zm] (Wm) {$W_m$} ; %
    \node[latent, above left=of Zm] (Um) {$U_m$} ; %
    \edge {Um} {Tm} ; %
    \edge {Um} {Zm} ; %
    \edge {Tm} {Ym} ; %
    \edge {Wm} {Zm} ; %
    \edge {Wm} {Ym} ; %
    \plate[inner sep=0.25cm, xshift=-0.12cm, yshift=0.12cm] {plate1}
    {(Tm) (Zm) (Ym)(Um)(Wm)} {m=1..M};
    \node[latent, above right=of Wm] (phi) {$\phi$};
    \node[latent, below left=of Tm] (lambda) {$\lambda$};
    \node[latent, below=of Ym] (Un) {$U_n$} ; %
    \node[obs, below right=of Un] (Zn) {$Z_n$} ; %
    \node[obs, below left=of Zn] (Tn) {$T_n$} ; %
    \node[latent, below right=of Zn] (Yn) {$Y_n$} ; %
    \node[latent, above right=of Zn] (Wn) {$W_n$} ; %
    \node[latent, above right=of Wn] (theta) {$\theta$};
    \node[latent, below left=of Un] (nu) {$\nu$};
    \node[latent, left=of Tm] (varphi) {$\varphi$};
    \edge {Un} {Zn} ; %
    \edge {Tn} {Yn} ; %
    \edge {Wn} {Zn} ; %
    \edge {Wn} {Yn} ; %
    \edge {theta} {Ym} ; %
    \edge {theta} {Yn} ; %
    \edge {phi} {Wn} ; %
    \edge {phi} {Wm} ; %
    \edge {lambda} {Un} ; %
    \edge {lambda} {Um} ; %
    \edge {nu} {Zn} ; %
    \edge {nu} {Zm} ; %
    \edge {varphi} {Tm} ; %
    \plate[inner sep=0.25cm, xshift=-0.12cm, yshift=0.12cm] {plate1}
    {(Tn) (Zn) (Yn)(Un)(Wn)} {n=M+1..N};
  }
  \caption{The M-bias Probabilistic Graphical Model.  The
    observational data is in the m=1..M plate where the intervention
    is given in M+1..N.  Parameters explicitly show how information
    transfers between the pre and post intervention world.}
  \label{mbias_full}  
\end{figure}

We now repeat the same analysis using Bayesian statistics and
probabilistic graphical models (PGM) instead of the do-calculus and causal
graphical models (CGM).  %The advantage of using this framework will%
                         %FIX here 
                         %be that we are able to leverage Bayesian statistics 
The appropriate graphical model is given in Figure
\ref{mbias_full}.  Some notable differences with the CGM structure are
that there are now two plates; one for the original graph (observations
$1..M$) and one for the mutilated graph (observations $N+1..M$).  As we
are now using Bayesian statistics every repeated observation has its
own index.  Another key difference is that arrow directions in the PGM
simply represent a factorization and not causality and can be reversed.
The inferential step uses standard Bayesian techniques
such as conditionalizing and marginalizing rather than the do-calculus.
 
The inference problem is to determine $Y_{1..N}$ for some hypothetical
$T_{N+1..M}$ and the data $Z_{N+1..M}, T_{1..M}, Y_{1..M}$.

The full probability specification is
\begin{align*}
P(&Y_{1..N}, W_{1..N}, Z_{1..N}, U_{1..N}, T_{1..M} ,\lambda, \phi,
    \theta, \varphi, \nu | T_{M+1, ..N}) = \\
& \left( \prod_{m=1}^M P(Y_m|W_m,T_m,\theta)P(T_M|U_m,\varphi) P(Z_m|U_m,W_m,\nu)
  P(U_m|\lambda) P(W_m|\phi) \right) \\
& \times \left( \prod_{n=M+1}^N P(Y_n|W_n,T_n,\theta)
  P(Z_n|U_n,W_n,\nu) 
  P(U_n|\lambda) P(W_n|\phi) \right) \\
& \times P(\lambda) p(\phi) P(\theta) P(\varphi) P(\nu).
\end{align*}

We note that there is no marginal distribution of $T_{M+1, ..N}$ where
in contrast $ T_{1..M}$ does indeed have a distribution and is
dependent on both $Z_{1..M}$ and hidden variables $ U_{1..M}$ and
$W_{1..M}$, which is precisely why it is hard to identify when the
outcome is \emph{caused} by the treatment and when the allocation of
treatments is associated with hidden variables that can predict the outcome.

Once we have the model specified in these terms we are able to predict
$Y_{M+1..N}$ conditional on the data and any counter factual $T_{M+1..N}$.
This simply takes the form:

\begin{align*}
&P(Y_{M+1..N} | W_{1..N}, Z_{1..N}, U_{1..N}, T_{1..M} , Y_{1..N},\lambda, \phi,
    \theta, \varphi,  T_{M+1, ..N}) \\
& = \frac{  \iiint P(Y_{1..N}, W_{1..N}, Z_{1..N}, U_{1..N}, T_{1..M} ,\lambda, \phi,
    \theta, \varphi | T_{M+1, ..N}) dW_{1..N} dU_{1..N} d\lambda d\phi
  d\theta d\varphi}
{  \iiint P(Y_{1..N}, W_{1..N}, Z_{1..N}, U_{1..N}, T_{1..M} ,\lambda, \phi,
    \theta, \varphi | T_{M+1, ..N}) dW_{1..N} dU_{1..N} d\lambda d\phi
  d\theta d\varphi dY_{M+1..N}}.
\end{align*}

\noindent
This is the most direct Bayesian solution to the M-bias problem as
posed by Pearl, but it turns out to be difficult computationally.
While Bayesian statistics often face high dimensional integrals 
similar to the above there are two issues that make this problem
relatively difficult.  The first of these issues is
that the role of latent variables in 
Pearl's causal graphical model framework and the probabilistic
graphical model framework popularized by Jordan
\cite{jordan2004graphical} and others is quite different. 
Pearl uses latent variables to represent any and all external complexities that
the world may impose on the system, therefore the latent variables
may have very large cardinality, in contrast, probabilistic graphical
models a latent variable (i.e. an unobserved variable within a plate)
is usually present to coerce the model into complete data exponential 
family form.  This enables families of approximating algorithms such as
Gibbs sampling \cite{geman1984stochastic} and mean field variational
Bayes \cite{ghahramani2001propagation}. 

The second issue is that this model has some parameters that are
identifiable and others that are not.  Indeed the point of this
exercise is that it is possible to infer treatment effects averaging
over $Z$, but it is not possible to infer personalized treatment
effects adjusting for $Z$.  This lack of identifiability means many
methods for locally  approximating the
posterior fail to represent the
identifiable and unidentifiable aspects of the posterior - which in
this case we \emph{are} interested in.

We handle these two issues by (a) re-parameterizing the model and
analytically marginalizing out $U_m$ (which allows $U_m$ to have
unbounded cardinality), (b) keeping the cardinality of $W_m$ to 2 for
demonstration purposes and (c)  using the MCMC algorithm known as the
``the independence sampler'', which while inefficient guarantees
exploration of the whole posterior\footnote{This algorithm becomes
  less efficient when there is more data, as this makes the posterior
  ``sharper'' and harder to hit the main support using such a naive
  proposal distribution, so we demonstrate on a modest sides datasets
  with large treatment effects.}.

Before proceeding it is worth mentioning another key difference between
causal graphical models and probabilistic graphical models.  In a
causal graphical model the direction of an arrow describes the
direction of causality, in contrast, a probabilistic graphical model
the arrow direction is just highlight one possible factorization which
can also be reversed i.e. using the identity
$P(a,b)=P(a|b)p(b)=P(b|a)P(a)$ and as such in the probabilistic
graphical model framework arrow reversals are permitted\footnote{The
  correct interpretation of a conditional probability $P(a|b)$ in Bayesian
  statistics and probabilistic graphical models is that learning $b$
  causes you to think certain values of $a$ are more or less likely,
  this point was highlighted by de Finetti when he said:
``I do not look for why THE FACT that I foresee will come about, but
why I DO foresee that the fact will come about. It is no longer the
facts that need causes; it is our thought that finds it convenient to
imagine causal relations to explain, connect and foresee the
facts. Only thus can science legitimate itself in the face of the
obvious objection that our spirit can only think its thoughts, can
only conceive its conceptions, can only reason its reasoning and
cannot encompass anything outside itself.'' \cite{de1989probabilism}.
This interpretation
is more or less faithfully implicit in  the probabilistic graphical
models and Bayesian statistics literature .  In the two plates
framework we are able to set up the model such that the conditional
probability is also meaningful causally, in the sense of if I set
$T_{M+1}$ it will cause me to think that $Y_{M+1}$ will have certain
values with higher or lower probability.  We can then if we choose do
a decision analysis where we have a utility function
$U(Y_{M+1},T_{M+1})$ and we choose $T_{M+1}$ to maximize this
utility.  A minor difference with how decision theory is usually
presented is that here the distribution of $Y_{M+1}$ changes with
$T_{M+1}$ and $U(Y_{M+1},T_{M+1})$ might often not depend on $T_{M+1}$
at all e.g. $U(Y_{M+1},T_{M+1})=Y_{M+1}$ is particularly common.}

%We use Bayes rule to reverse arrows and the sum rule to produce
%marginalizations in order to put the observational data in the $m$
%plate into as similar as possible form to the intervention plate $n$.

%We suggest these heuristics for re-parameterization the plates:

%\begin{enumerate}
%\item
%The two plates should have the same arrow directions where possible,
%this allows parameters to be shared and for information to flow in a
%transparent way from the observation plate to the intervention plate.
%\item
%The outcome variable should be a conditional probability in both
%plates, the motivation is simply is this is a primitive in determining
%treatment effects.
%\item
%The treatment variable should be conditioned upon in both plates, this
%is really an extension of (1) as the treatment variable by definition
%is conditioned on in the second graph.  It is usually a good idea to
%condition on covariates in both plates too.
%\item
%Subject to the parameters being shared between the two plates, all
%latent variables should be marginalized analytically, be aware that
%marginalizations can cause prior distributions between the two plates
%to have joint distributions.
%\end{enumerate}

We see straight away that $P(Y|W,Z,\theta)$ is already in a convenient
form in both plates.  However we have some work to do in simplifying
$P(W_m,U_m,Z_m,T_m|\varphi,\lambda,\phi,\nu)$ and
$P(W_n,U_n,Z_n|T_n,\lambda,\phi,\nu)$ in particular factorizing
$P(W_m,U_m,Z_m,T_m|\varphi,\lambda,\phi,\nu)$ and marginalizing $U_m$
(it turns out $W_m$ cannot be marginalized without causing the two
plates to have different structure: 

\begin{align*}
P&(W_m,|Z_m,T_m,\varphi,\lambda,\phi,\nu) \\
& = \frac{ \int P(W_m,U_m,Z_m,T_m|\varphi,\lambda,\phi,\nu) dU_m  }
{ \iint P(W_m,U_m,Z_m,T_m|\varphi,\lambda,\phi,\nu) dU_m dW_m }  = P(W_m|Z_m,T_m,\alpha).
\end{align*}

\noindent
where $P(W_m|Z_m,T_m,\alpha)$ is a re-parameterization integrating over
the (possibly) high cardinality $U_m$ and $\alpha$ is a transform of
the parameters $\varphi,\lambda,\phi,\nu$.  In the intervention plate a
different computation occurs as knowledge of $T_n$ no longer gives any
information about $W_n$ (this occurs because we set $T_n$ ourselves).

\begin{align*}
P&(W_n,|Z_n,\lambda,\phi,\nu)  = \frac{ \int P(W_n,U_n,Z_n|\lambda,\phi,\nu) dU_m  } 
{ \iint P(W_n,U_n,Z_n|\lambda,\phi,\nu) dU_m dW_m }  = P(W_m|Z_m,\omega)
\end{align*}

\noindent
where $P(W_m|Z_m,\omega)$ is a re-parameterization integrating over
the (possibly) high cardinality $U_m$ and $\omega$ is a transform of
the parameters $\lambda,\phi,\nu$. 

We can also do a similar
re-parameterization for $P(T_m|\kappa)$ and $P(Z|\nu)$ although these quantities are
ancillary to the analysis.

Details for the transforms for $\alpha$, $\omega$ in terms of the
original parameters are given in the supplementary material.

As $\alpha$ and $\omega$ are both functions of $\lambda,\phi,\nu$ the
prior distribution has a dependency i.e. $P(\alpha,\omega) =
P(\omega|\alpha)P(\alpha)$, further note that $\alpha$ encodes a
distribution that conditions on $T$ where $\omega$ does not as such
$\alpha$ is typically of higher dimension than $\omega$.  The new
parameterization the graph is given in Figure \ref{mbias_reversal}. 

We briefly comment on the reversal of the arrow direction from the
treatment which has the interpretation of $T_m$ causes you to think
that $W_m$ may have certain values, but which is not permitted in a
causal graphical model as $W_m$ causes $T_m$ and the arrow cannot be
reversed as the arrow has a causal interpretation.  The model
can now be written:

\begin{figure}
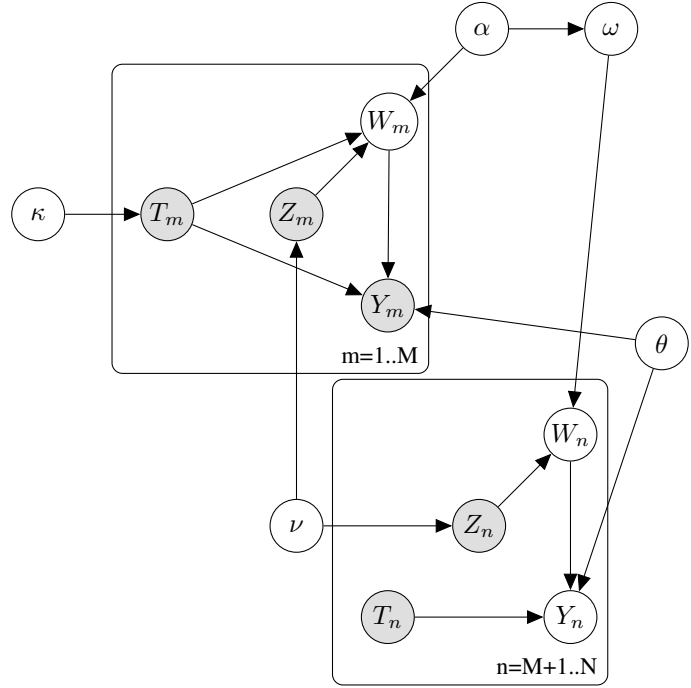

  \centering
  \tikz{ %
    \node[obs] (Zm) {$Z_m$} ; %
    \node[obs, left=of Zm] (Tm) {$T_m$} ; %
    \node[obs, below right=of Zm] (Ym) {$Y_m$} ; %
    \node[latent, above right=of Zm] (Wm) {$W_m$} ; %
    \edge {Tm} {Ym} ; %
    \edge {Tm} {Wm} ; %
    \edge {Zm} {Wm} ; %
    \edge {Wm} {Ym} ; %
    \plate[inner sep=0.25cm, xshift=-0.12cm, yshift=0.12cm] {plate1}
    {(Tm) (Zm) (Ym)(Um)(Wm)} {m=1..M};
    \node[latent, above right=of Wm] (alpha) {$\alpha$};
    \node[latent, right=of alpha] (omega) {$\omega$};
    \node[obs, below right=of Un] (Zn) {$Z_n$} ; %
    \node[obs, below left=of Zn] (Tn) {$T_n$} ; %
    \node[latent, below right=of Zn] (Yn) {$Y_n$} ; %
    \node[latent, above right=of Zn] (Wn) {$W_n$} ; %
    \node[latent, above right=of Wn] (theta) {$\theta$};
    \node[latent, below left=of Un] (nu) {$\nu$};
    \node[latent, left=of Tm] (kappa) {$\kappa$};
    \edge {Tn} {Yn} ; %
    \edge {Zn} {Wn} ; %
    \edge {Wn} {Yn} ; %
    \edge {theta} {Ym} ; %
    \edge {theta} {Yn} ; %
    \edge {omega} {Wn} ; %
    \edge {alpha} {Wm} ; %
    \edge {nu} {Zn} ; %
    \edge {nu} {Zm} ; %
    \edge {kappa} {Tm} ; %
    \edge {alpha} {omega} ; %
    \plate[inner sep=0.25cm, xshift=-0.12cm, yshift=0.12cm] {plate1}
    {(Tn) (Zn) (Yn)(Un)(Wn)} {n=M+1..N};
  }
  \caption{The M-bias Graph Probabilistic Graphical Model with arrow Reversals (and $U$ marginalized}
  \label{mbias_reversal}  
\end{figure}

\begin{align*}
P&(Y_{1..N}, W_{1..N}, Z_{1..N}, \theta, \alpha, \omega, \nu | T_{1..N}) = \\
&\left( \prod_{m=1}^M P(Y_m|W_m,Z_m,\theta) P(W_m|Z_m,T_m,\alpha)
  P(Z_m|\nu) \right) \\
&\times \left( \prod_{n=N+1}^N P(Y_n|W_n,Z_n,\theta) P(W_n|Z_n,\omega)
  P(Z_n|\nu) \right) P(\theta) P(\alpha)P(\nu) P(\omega|\alpha)
\end{align*}

\noindent
The M-structure prevents the identification of individual treatment
effects, yet average treatment effects can be inferred.
The identification of $\theta$ is hampered by it involving the latent
$W_m$, similarly the identification of $\alpha$ is also hampered by the
unobserved latent variable, but there is the further problem that
$\alpha$ is present in the observation plate, but $\omega$ is in the
intervention plate and only $P(\omega|\alpha)$ can transfer
information between the plates, and we may contemplate the situation
where $P(\omega|\alpha) P(\alpha) =P(\omega)P(\alpha)$ where information simply
doesn't flow.

It is remarkable that in the face of all this poor identifiability
that we can recover the M-Bias result found by Pearl.  That is we can
infer the average treatment effects if not personalized treatments
adjusting for $Z$.

We now specify parametric forms for the various terms in the model:
\begin{center}
\begin{tabular}{cc}
$P(Y_m=1|T_m,W_m,\theta) = \theta_{T_m,W_m}$ & $P(W_m|Z_m,T_m,\alpha)= \alpha_{W_m,Z_m,T_m}$ \\
$P(W_n|Z_n,\omega) = \omega_{W_n,Z_n}$ & $P(Z_m|\nu) = \nu_{Z_m}$\\
$\forall z \hspace{4mm} \omega_{:,z} \sim {\rm Dirichlet}(1_K)$ & $\forall z,t \hspace{4mm} \alpha_{:,z,t} \sim {\rm Dirichlet}(1_K)$\\
$\forall w,z \hspace{4mm} \theta_{w,z} \sim {\rm Beta}(1,1)$& $\nu \sim {\rm Dirichlet}(1_K)$\\
\end{tabular}
\end{center}

We also define:

\[
\psi_{z,t} = \sum_{w'} \theta_{t,w'} \alpha_{w',z,t}.
\]

\noindent
Which has the interpretation

\[
P(Y_m=1|Z_m,T_m) = \psi_{Z_m,T_m}.
\]

\noindent
The transformed parameter $\psi$ unlike $\theta, \alpha$ is identifiable, but has
no causal interpretation (rather it could be used for predicting a
missing $Y_m$ entry in the first plate only).  Using $\psi$ we are
able to write the log likelihood function for $\omega, \alpha$ (we can
infer $\nu$ separately):

\noindent
Finally we also define:

\[
\rho_{z,t} = \sum_{w'} \theta_{t,w'} \omega_{w',z}.
\]

\noindent
Which has the interpretation

\[
P(Y_n=1|Z_n,T_n) = \rho_{Z_n,T_n} .
\]

\noindent
The transformed parameter $\rho$ is not identifiable, but does have a
causal interpretation.  We can now write the likelihood:

\[
P(Y_{1..M}|X_{1..M},T_{1..M},\psi) = \mathcal{L}(\psi) = \sum_m^M Y_m \log(\psi_{Z_m,T_m} ) + (1-Y_m) \log(1-\psi_{Z_m,T_m} ).
\]

\section{Simulation Study on Example using MCMC}
\label{ssmcmc}
A simple inference algorithm is to sample a proposal of the parameters
$(\omega^*,\alpha^*,\theta^*)$ from the
prior, compute $\psi^*$ as a function of the parameters and then
evaluate the likelihood to weight the samples i.e. an importance
sampling algorithm.  A slight variant of the importance sampling
algorithm is the MCMC algorithm known as the independence sampler
\cite{gamerman2006markov}.  In
this case the proposal distribution is again the prior distribution
but a transition is accepted with probability $\mathcal{L}(\psi^*) +
\log P (\omega^*,\alpha^*,\theta^*) -
\mathcal{L}(\psi) -\log P (\omega,\alpha,\theta) > \log(u)$, where $u \sim {\rm U}(0,1)$.  This algorithm
is not efficient, as it does not concentrate exploration on good parts
of the posterior however this lack of state is ideal in this case
where we have a complex posterior with multiple isolated modes (high
posterior regions that are far from each other and therefore difficult
to approximate).  Markov chain Monte Carlo
algorithms such as Gibbs sampling and Hamiltonian Monte Carlo get caught in these
isolated modes.
This algorithm is ideal for a simple situation such as this but will scale
very poorly for more complex problems (or large data sets).  Some
sophisticated Monte Carlo
methods have an improved ability to escape
isolated modes e.g. \cite{neal2001annealed}
\cite{betancourt2014adiabatic}.

\begin{figure}%
    \centering
    \subfloat[Personalized Treatment Effect Adjusted for $W$]{{\includegraphics[width=6cm]{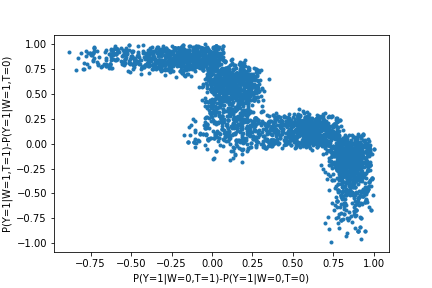}}}%
    \qquad
    \subfloat[Personalized Treatment Effect Adjusted for $Z$]{{\includegraphics[width=6cm]{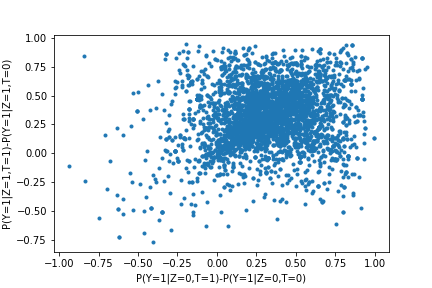} }}%
    \qquad
    \subfloat[Average Treatment Effect
    ]{{\includegraphics[width=6cm]{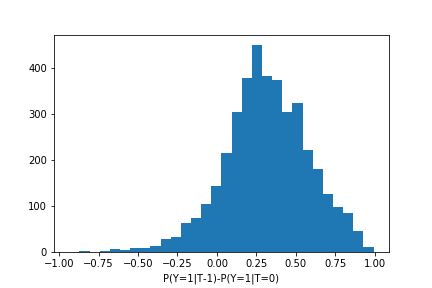} }}% 
    \caption{Posterior samples of the treatment effect, adjusted for
      $W$, adjusted for $Z$ and averaged}%
    \label{posterior_treat_cond}%
\end{figure}

We might also be interested in plotting $P(W_n|Z_n)$ but we can note that
this is, in this case, completely determined by the prior distribution.
As we assume independence i.e. $P(\omega|\alpha)=P(\omega)$, so the
posterior of $\omega$ is the same as the prior.

After a burn in of $50000$, we draw $2 \times 10^8$ MCMC samples but thin them by only retaining every
$50000$th sample giving $4000$ samples.  In Figure \ref{posterior_treat_cond} (a) the treatment effect
conditional on the unobserved $W$ is shown, where in Figure
\ref{posterior_treat_cond} (b) the treatment effect 
conditional on the observed $Z$ is shown.   

As expected Figure \ref{posterior_treat_cond} (b) shows that we are
unable to determine the personalized treatment effect adjusting for
$Z$.  The posterior gives support to many possible values and has
little structure other than there is some posterior correlation
i.e. the treatment effect for $Z=0$ are somewhat more likely to be
similar to $Z=1$.

It is also instructive to look at the treatment effects adjusting for
the unobservable $W$ in Figure \ref{posterior_treat_cond} (a), here
there is also a lot of uncertainty but 
there is also some notable structure in that there is an
anti-correlation between the treatment effects when $W=0$ and when
$W=1$.  The ``M'' shape in the posterior reveals a surprising
structure in the posterior, we do not have a complete adequate
explanation for this currently and intend to investigate it further.

This anti-correlation is critical to being able to determine the
non-personalized treatment effects.  This is due to that
unconditionally: $P(W=0)=P(W=1)=0.5$ as such the non-personalized
treatment effect:

\begin{align*}
&P(Y_{m+1}=1|T_{m+1}=1) - P(Y_{m+1}=1|T_{m+1}=0) \\
& = \left(P(Y_{m+1}=1|T_{m+1}=1,W_{M+1}=0) -
  P(Y_{m+1}=1|T_{m+1}=0,W_{M+1}=0) \right) P(W_{M+1}=0)\\
& + \left(P(Y_{m+1}=1|T_{m+1}=1,W_{M+1}=1) -
  P(Y_{m+1}=1|T_{m+1}=0,W_{M+1}=1) \right) P(W_{M+1}=1),
\end{align*}

which we can interpret as (half) the sum of the $x$ and $y$ axis in
Figure \ref{posterior_treat_cond} (a), which we can see by eye has
much less variance, i.e. when $P(Y=1|W=0,T=1) - P(Y=1|W=0,T=0)$ is
high, then $P(Y=1|W=1,T=1) - P(Y=1|W=1,T=0)$ is low and
correspondingly the sum of the two is stable.  The posterior of
the average 
treatment effect is given in Figure \ref{posterior_treat_cond} (c).

\section{Conclusion}
\label{conc}
In this paper it was shown that if we follow Rubin's advice to condition on all
variables we recover Pearl's result that under M-structures average treatment
effects are identifiable, yet personalized treatment effects are not.
The posterior samples revealed a surprising ``M'' shape which we do
not have an adequate explanation of at this time, but intend to
investigate further.

The methodology we use is standard Bayesian statistics 
using a two plate probabilistic graphical model where one plate
represents the observational data and the other the post-intervention
data.  The two plates could be seen as analogous to the pre and post
mutilation graphs in the CGM paradigm. 

The two plates framework appears interesting as a general tool for
casting causal problems usually analyzed using the do-calculus in a
framework that can be analyzed using the vast tools of Bayesian
statistics and which includes a methodology that is correct for finite
samples and by the use of prior distributions can even draw inference
in cases that would be deemed non-identifiable by the do-calculus.  It
is conjectured that the two plates framework can be proven to recover
the identifiability results of the do-calculus.

\section{Acknowledgments}

The author is grateful for comments on an earlier draft from 
Noureddine El Karoui, Jeremie Mary, Mike Gartrell and Flavian Vasile
which improved the paper.  I am especially grateful to Finn Lattimore
for many interesting discussions and for every time our intuition
disagreed to be another learning opportunity.

\bibliographystyle{plain}
\bibliography{literature}

\end{document}